\RequirePackage{fix-cm}
\documentclass[twocolumn]{svjour3}          % twocolumn
\smartqed  % flush right qed marks, e.g. at end of proof
\usepackage{amssymb}
\usepackage{graphicx}
\usepackage{natbib}
\usepackage{url}
\usepackage{amsmath}
\usepackage[ruled,vlined]{algorithm2e}
\usepackage{caption}
\usepackage{booktabs}
\usepackage{float}

\usepackage{xcolor}

\usepackage[font=small,labelfont=bf]{caption}
\journalname{(Under revision)}
\begin{document}
%\linenumbers
\title{Interpretable reinforcement learning with decision-tree pruning}

% What would you think of "Post-Hoc Pruning of Program-Extracted Policies: Non-Inferiority Simplification for Interpretable RL." ? which would better reflect the process and application area?



\author{Mark Ringer       
    \and
        Michel Tokic 
}
\institute{M. Ringer \at
Faculty of Mathematics, Informatics and Statistics, \\Ludwig-Maximilians-University Munich, Munich, Germany, \\
mark.leon.ringer@gmail.com
\and
M. Tokic \at
Siemens AG, Data \& Artificial Intelligence, Otto-Hahn-Ring~6, 81739~Munich, Germany \\
and \\
Faculty of Mathematics, Informatics and Statistics, \\Ludwig-Maximilians-University Munich, Munich, Germany,
michel@tokic.com
}

\date{Received: 4th May 2026 / Accepted: Under revision}

\maketitle

\begingroup
\sloppy
\begin{abstract}
Reinforcement learning policies are difficult to inspect, but interpreting them is a prerequisite for trustworthiness. Converting a trained policy into explicit decision-tree rules improves transparency and the resulting artifacts often remain too complex for human understanding. We present a pruning process that simplifies such 
rule-based policies while preserving task performance and making edits to the policy auditable. The process defines a small set of structural and usage-aware operators and evaluates candidate edits by 
re-executing the policy to measure return and interpretability proxies. This exposes an transformation process from complex to compact policy structures. We investigate this approach on classic control and MuJoCo benchmarks, where pruning traces reveal consistent interpretability improvements while maintaining high performance.

\end{abstract}
\endgroup

%Do we cite in the abstract? 
\section{Introduction}
\label{intro}
Reinforcement learning has reached, and in some cases exceeded, human-level performance in control and games \citep{silver2017,Vinyals2019}, but the resulting policies are often hard to inspect \citep{Henderson_2018,zahavy16}. In settings where verification and accountability are required \citep{doshivelez2017,lipton2017}, intransparent policies, e.g. represented as neural networks, are typically not an option. Program-extracted policies address part of this gap by transforming trained neural network actors into explicit decision trees \citep{verma2019,bastani2019v,delfosse2023Interpretable}. These artifacts are executable and analyzable, but they often remain too large for reliable human understanding. Interpretable policies must be compact enough to read and simulate, not only explicit in form.

This paper studies pruning as a principled, post-hoc simplification operator on already interpretable, program-extracted policies. We treat simplification as a controlled edit process: apply a candidate operator, re-execute the policy to measure task return and interpretability proxies, and accept the edit only if it passes a non-inferiority test. Each accepted edit is recorded, yielding an auditable trail that ties structural changes to measured effects. Interpretability is thus a property of the transformation trajectory as well as of the final program.

\section{Method}
\label{sec:method}
This sections details the transformation process from common RL policies represented as neural networks, particularly we used the actor network from stable-baselines3 \citep{stable-baselines3}, into decision trees using the transformation method proposed by \cite{kohler2024distilling}. 
We then employ a lightweight benchmark to evaluate the obtained policies, inspired by \cite{kohler2025interpretable}. The same benchmark is reused to assess our pruned policies under identical conditions.

\subsection{Distilling neural network actors into decision-tree rules} \label{subsec:distill}
The first step of the process is to transform a neural network policy into a  decision tree as  described in \citep{kohler2024distilling}. Here, the teacher policy, in our case a neural network actor from from stable baselines, is used to generate a large corpus of environment specific state-action pairs. These pairs are then used to fit a scikit \textit{DecisionTreeClassifier} \citep{scikit-learn}, called the learner.

\subsection{Improving Interpretability}
In general, the interpretability of decision trees can be improved through pruning strategies. A common way to quantify interpretability is by measuring the number of resulting leaf nodes.

\paragraph{Leaf-node-based interpretability proxy}
We use the number of leaf nodes as a proxy for interpretability. This measure captures the complexity of a given function while remaining invariant to syntactic variations.

\paragraph{Pruning strategies}
In extending \citep{kohler2025interpretable}, our objective is to advance interpretability by focusing exclusively on Python policy trees and developing a structured pruning framework.
Decision trees have consistently been found to be more interpretable than neural policies in human-subject studies \citep{freitas2014, lipton2017}, primarily because their hierarchical and rule-based structure aligns with human reasoning. While Kohler’s framework treated trees and MLPs as equally valid policy classes, we deliberately restrict our investigation to tree-based policies to maximize interpretability and enable direct structural analysis.

To enhance interpretability in decision trees, we apply systematic pruning to distilled policies.
Each tree is iteratively reduced using  one of the following three  strategies defined in Sec.~\ref{sec:pruning-algos}:

\begin{itemize}
    \item Max-depth pruning 
    \item Max-impurity pruning 
    \item Decision-tree Adaptive Constrained Pruning 
\end{itemize}

Each structural pruning step is followed by a subtree collapsing pass in order to remove residual redundancies and compress the tree further.

%In parallel, we visualize all intermediate tree versions to enable direct interpretability assessment (e.g. see Figure \ref{fig:policy-comparison}).
Visual inspection of consecutive trees makes it possible to track which branches are pruned, whether these changes affects decision logic, and how they correlate with benchmark performance.
By linking each pruning step to its corresponding performance measurement, we gain fine-grained insight into the trade-off between simplicity and reward.

The resulting policies are not only quantitatively interpretable, according to Kohler’s proxies of smaller size, but also qualitatively transparent, as their structural evolution can be directly observed. In contrast to Kohler’s approach, which yields compact policies through direct training, our method makes the simplification process itself explicit, revealing how complex trees are incrementally transformed into minimal forms. This transparency enhances the interpretability of both the final policies and the underlying transformation.

\subsection{Description of Pruning Algorithms}
\label{sec:pruning-algos}

We chose pruning algorithms to optimize the initial policy in terms of interpretability. \textit{Max-depth} plus subtree collapsing is a structure-only cut followed by deterministic redundancy cleanup. \textit{Max-impurity} plus subtree collapsing adds local data statistics (node purity) to make targeted reductions. \textit{Decision-tree Adaptive Constrained Pruning} (DACP) incorporates runtime usage, in form of node visit counts, with reward guards.

\paragraph{Subtree collapsing}

\label{subsec:uniform-pruning}
We simplify the decision trees by recursively collapsing decision nodes into leaf nodes if both childern are leafs and contain the same action \citep{costcomplexity}. We use this method to support the other strategies but don't treat it as a complete strategy.
The behavior of forming subtrees with identical actions could occur because the distilling algorithm used in \citep{kohler2024distilling} does not punish unnecessary splits, or later during pruning when alternative actions are removed. 
This operation does not alter the predictions of the model but simplifies the tree considerably. 
By eliminating such uniform subtrees, the resulting model becomes smaller, easier to interpret, and more efficient to evaluate, while maintaining identical predictive behavior.

\subsubsection{Max-Depth Pruning}
\label{subsec:max-depth-pruning}
Max-depth pruning restricts the maximum depth of the decision tree to a predefined limit. 
During training or post-processing, any node that would extend beyond this depth is replaced by a leaf node. 
The leaf represents the majority class of the samples that fall into it, effectively summarizing the deeper part of the tree. 
This approach prevents the model from growing overly complex and helps control overfitting by enforcing a global constraint on tree size and depth \citep{maxdepth}. 
In our case, we prune the tree in steps by reducing the max depth in discrete steps. We always prune the original tree, the steps are saved to produce a trajectory to find an optimal policy. After each max depth pruning step, uniform pruning is applied.

\subsubsection{Max-Impurity Pruning}
\label{subsec:max-impurity-pruning} 
Max-impurity pruning reduces the tree’s complexity by halting further splits once a node becomes sufficiently homogeneous with respect to its class distribution. 
The impurity of a node is computed from the distribution of classes using the following metric based on the Gini index, 
\[
\text{Max-Impurity} = 1 - \sum_{k=1}^{K} p_k^2,
\]
where $p_k$ denotes the proportion of samples belonging to class $k$ in that node and $K$ denotes the set of classes \citep{costcomplexity}. 
Nodes with low impurity, meaning that one class dominates the samples within them, are considered pure enough and are converted into leaf nodes. 
This pruning strategy prevents unnecessary splits in already homogeneous regions of the data, thereby simplifying the tree and often improving its generalization performance.
After max-impur ity pruning we apply uniform pruning to simplify the tree.

\subsubsection{\textbf{D}ecision-tree \textbf{A}daptive \textbf{C}onstrained \textbf{P}runing }
\label{subsec:pact}

This pruning method, algorithm \ref{alg:heuristic_pruning} in the appendix, is inspired by the A* Algorithm \citep{astar}, Cost-Complexity Pruning \citep{costcomplexity} and Critical Value Pruning \citep{criticalvalue} and based on the assumption that nodes that are less important for the performance of a given policy tree are visited less often than more critical nodes. The idea of heuristic pruning has previously been explored for different problems, for example in \citep{feldotto2022network} heuristic network pruning, where a similar concept for neural networks is explored. A similar method is also used by humans do to limit cognitive resources if tasked with evaluating probabilistic planning tasks \citep{sass2025heuristic}.

Following the intuition that this is also applicable for RL policies in the form of decision trees, a given policy $\pi$ can be pruned by counting, for each node $n$, how often it is visited $\mathbf{C_i} = (c_{i,1}, c_{i,2},...,c_{i,N})$ and removing the $k$ nodes with the lowest visit count. This process can be repeated incrementally until the desired policy size is achieved. Obtaining the visit counts is accomplished by by using a $counters(\pi)$ function that runs the policy and collects for each node how often it was visited.

However, a challenge with this approach is, that some nodes with a low visit count ($c_i$) are actually crucial to maintain a high reward. 
Therefore, it is necessary to ensure that the reward doesn't decrease too fast by comparing the performance measured as cumulative reward $R_{i}$ of the pruned policy $\pi_{i}$ to the performance $R_{i-1}$ of its predecessor $\pi_{i-1}$. To formalize this, we define two parameters where $R_{Base}$ is the average reward of the original policy.:

\begin{itemize}
\item \emph{tolerance factor} $\delta \in (0,1)$ which controls the maximum allowed reward decrease per iteration
\[
\Delta = |R_{Base}| \times \delta 
\]
\item  \emph{stability factor} $\phi \in (0, 1 - \delta)$, which determines the absolute minimum reward that is considered acceptable
\[
\Phi = R_{Base} - |R_{Base}| \times (1 - \phi)
\]
\end{itemize}
where $\Delta$ is the maximum absolute decrease in reward that is still accepted and $\Phi$ is the minimal reward that is still accepted. 
To prevent the required reward from increasing beyond a reasonable level, we also defined a ceiling value $\Gamma$ as:
\[
\Gamma = R_{Base} - \Delta.
\]
Using these definitions, the minimum acceptable reward for each iteration $i$ is given by
\begin{equation}
R_{\min,i} = \min(\max(R_{i-1}-\Delta, \Phi), \Gamma).
\end{equation}
A pruning step can only be considered successful if the resulting policy $\pi_i$ achieves a reward $R_i \geq R_{\min, i}$.

Determining the reward of a policy requires a reliable benchmarking:
\\
$
R_i =  benchmark(\pi_i).
$
\\
This process is computationally expensive, and therefore it is necessary to minimize the number of $benchmark$ calls. 

Therefore, the pruning is performed on a batches of $k$ candidate nodes. To allow divide and conquer, $k$ fulfills $k \in \{ 2^m \mid m \in \mathbb{N}_{0}\}$. If the pruning for a batch is rejected, the candidate nodes are split in half and the pruning is recursively repeated. In this way, only those nodes of a batch whose removal causes $R_i < R_{\min,i}$ are not removed. 

If $|\pi'| = |\pi|$, we assume all nodes in the batch are essential and treat them as failed. Otherwise, we add all newly failed nodes to the failed list, clean the tree by using our $collaps$ function as described in section \ref{subsec:uniform-pruning}, collect new $counters$ and recalculate the batch size if there were any failed nodes.

If its no longer possible to remove candidates, we start trying to remove nodes that failed in previous steps. After we cannot remove any node anymore without the reward falling below the threshold, the algorithm terminates.

\section{Results}

\begin{table*}[h!]
\centering
\begin{tabular}{llcc}

\toprule
Algorithm & Agent Version (Teacher) & Learner Reward & Teacher Reward \\
\midrule
PPO & ppo-Acrobot-v1 & -86 $\pm$ 44 & -84 $\pm$ 25\\
PPO & ppo-CartPole-v1 & 488 $\pm$ 59 & 500 $\pm$ 0 \\
SAC & sac-HalfCheetah-v3 & 5023 $\pm$ 356 & 8898 $\pm$ 124 \\
PPO & ppo-LunarLander-v2 & 233 $\pm$ 58 & 149$\pm$ 34\\
SAC & sac-LunarLanderContinuous-v2 & 236 $\pm$ 96 & 262 $\pm$ 64\\
DQN & dqn-MountainCar-v0 & -102 $\pm$ 11 & -101 $\pm$ 10\\
SAC & sac-MountainCarContinuous-v0 & 94 $\pm$ 2 & 94 $\pm$ 2 \\
PPO & ppo-Pendulum-v1 & -207 $\pm$ 212 & -174 $\pm$ 107 \\
TD3 & td3-Swimmer-v3 & 355 $\pm$ 2 & 355
$\pm$ 2\\
SAC & sac-Walker2d-v3 & 1815 $\pm$ 1061 & 3917 $\pm$ 401 \\

\bottomrule
\end{tabular}
\caption{Performance comparision of original (Teacher) and distilled (Learner) policies. Note: The term "Learner" refers to the  policy after transforming the neural network Teacher policy into the initial decision tree using approach of \cite{kohler2024distilling}, which will be further optimized as depicted in Fig.~\ref{fig:performance}.}
\label{tab:distilled}
\end{table*}

\label{sec:results}
\begin{figure*}[h!]
    \centering
    \includegraphics[width=1\linewidth]{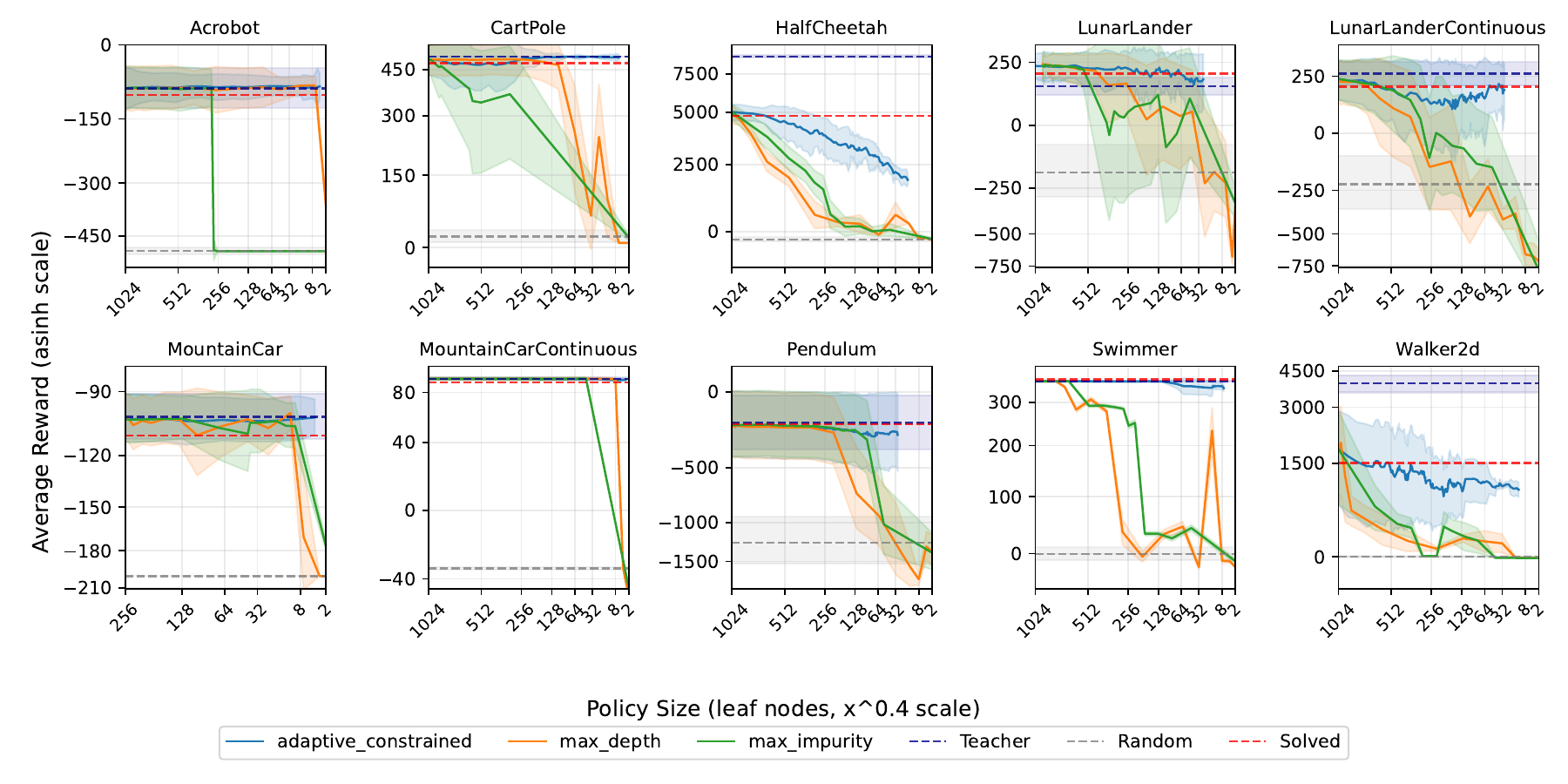}
    \caption{Performance of pruned policies, policy size (leaf nodes) against episodic reward. The teacher threshold marks the performance of the original model, the solved threshold marks the performance at which the environment is considered to be solved and the random threshold shows the performance of an agent taking a random action. For both the environments Pendulum and Walker2d there is no official solved threshold, so we defined solved for Pendulum as -200 and for Walker2d as 1500. }
    \label{fig:performance}
\end{figure*}

\begin{figure}[t]
    \centering
    \includegraphics[width=\columnwidth]{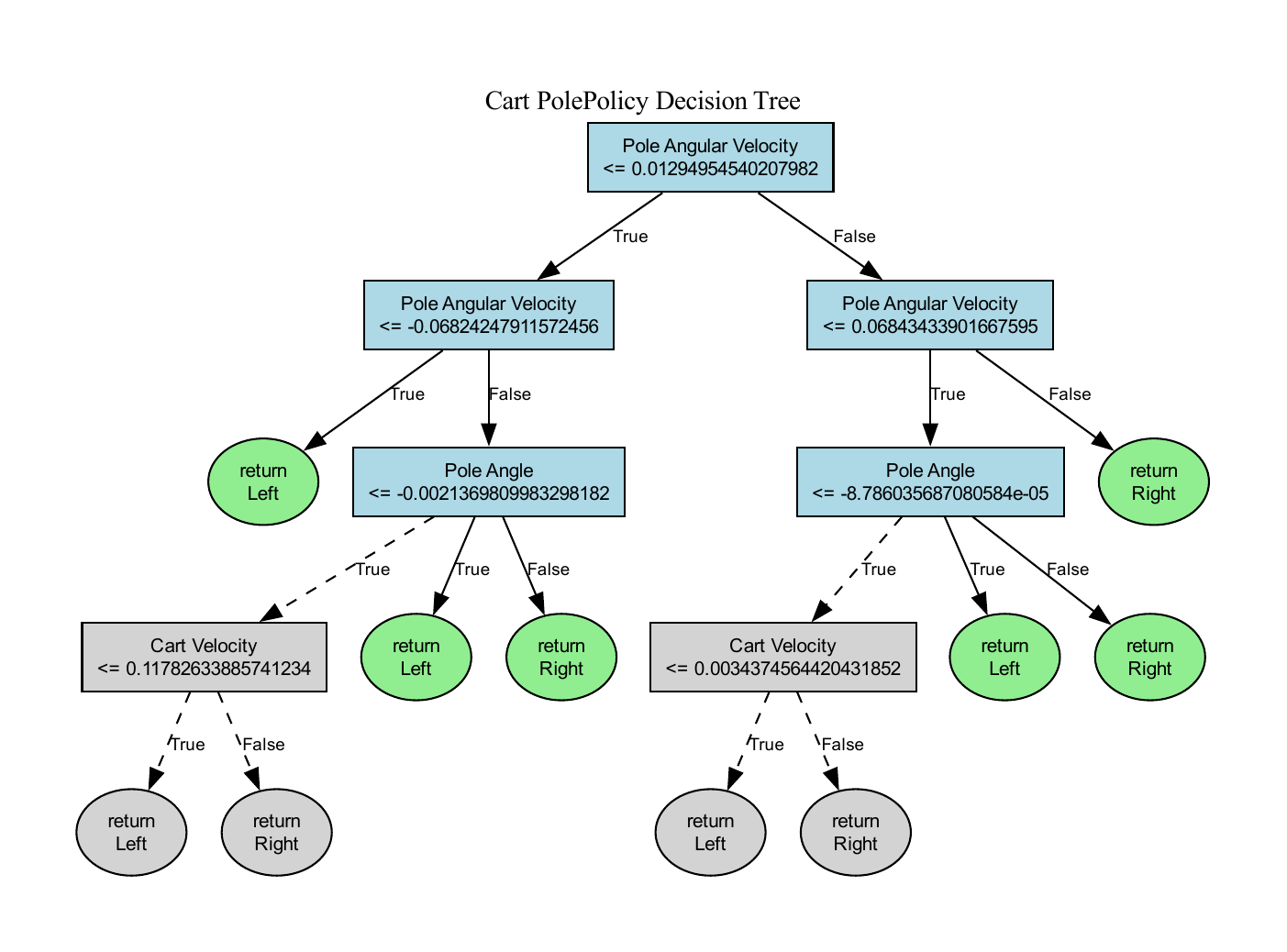}
    \caption{Last step of pruning the CartPole policy with DACP with a fixed batch size of 4. Here, the gray branches will be removed in the next iterative step, reducing the policy from eight to six leaf nodes while maintaining the performance (cf. Fig.~\ref{fig:performance}, adaptive\_constrained in CartPole-v1).}
    \label{fig:cartpole}
\end{figure}

\subsection{Distilling Policies}
We configured the pruning algorithm to allow a maximum number of up to 1024 leaf nodes. 
%This number was chosen to allow a sufficently complex starting policy while maintain a managable policy size. 
For simple tasks like CartPole, a much smaller tree would have been sufficient, nonetheless the tree grows to a much larger of leaf nodes if allowed. For more complex tasks such as Walker2D a larger number of leaf nodes could still yield improved performance, but to have a comparable starting state we decided to allow all trees to grow up to the same maximum size. Notably, the policy trees for MountainCar never increased in number of leaf nodes beyond 340 for any configured max leaf nodes, up to 2048. 

Table~\ref{tab:distilled} shows the performance of the distilled policies compared to their respective teachers. It can be observed that for most environments the performance of learner and teacher is similar. For complex tasks such as HalfCheetah and Walker2D, the episodic reward of the learner is much lower, probably due to the limited number of leaf nodes in the initial policy transformation process (neural network $\rightarrow$ decision tree). 
%\red{pendulum solved is max\_steps, also check for walker} 
Another noteworthy observation is that for LunarLander the teacher performed much worse than the learner. This could possibly be due to an overfitted teacher, reducing complexity turned out to  generalize better.

\subsubsection{Solved thresholds}
Whenever possible, we use the solved thresholds provided by the gymnasium library \cite{towers2024gymnasium}. 
Since gymnasium does not provide solved thresholds for Pendulum and Walker2d, we defined them as -200 for Pendulum and 1500 for Walker2d. For Pendulum, the optimal reward is 0 since the reward is calculated using
\[
r = -\left(\theta^2 + 0.1 \times \dot{\theta}^2 + 0.001 \times \text{torque}^2\right).
\] where $\theta$ is the pendulums angle, normalized between $[-\pi, \pi]$ while $torque$ represents the action space of the environment \cite{towers2024gymnasium}. A near-optimal performance where only small deviations from upright position occur could be defined by accumulating on average no more than -1 reward per time step. For a episode length of 200, we therefore define the solved threshold as -200.
For Walker2d 1500 is derived from the decomposed reward structure:
\begin{itemize}
\item healthy\_reward: Every timestep that the Walker2d is alive, +1 as reward.
\item forward\_reward: A reward for moving forward, depending on the velocity.
\item ctrl\_cost: A negative reward for taking large actions.
\end{itemize}
\cite{towers2024gymnasium}. To achieve an average reward of 1500 or higher, constant locomotion in the target direction with and average of 0.5 in rewards is required. Therefore, a reward of 1500 implies constant and stable locomotion in the right direction. Since the theoretical reward for movement has no upper bound, this threshold should be taken with caution.
\\
\begin{table}
\centering
\label{tab:lunar_lander_heuristic_subset}
\begin{tabular}{ccc}

\toprule
Leaves & Reward & Teacher \% \\
\midrule
1024 & 236.5 $\pm$ 95.5 & 89.8\% \\
887 & 226.9 $\pm$ 96.2 & 86.2\% \\
769 & 230.6 $\pm$ 110.2 & 87.6\% \\
713 & 218.1 $\pm$ 120.1 & 82.8\% \\
654 & 219.0 $\pm$ 125.5 & 83.2\% \\
$\dots$ & $\dots$ & $\dots$ \\
126 & 163.5 $\pm$ 149.8 & 62.1\% \\
121 & 136.8 $\pm$ 157.8 & 51.9\% \\
120 & 154.6 $\pm$ 146.3 & 58.7\% \\
117 & 149.0 $\pm$ 162.1 & 56.6\% \\
116 & 137.9 $\pm$ 164.4 & 52.4\% \\
$\dots$ & $\dots$ & $\dots$ \\
33 & 194.1 $\pm$ 119.0 & 73.7\% \\
32 & 206.4 $\pm$ 91.0 & 78.4\% \\
31 & 168.2 $\pm$ 157.2 & 63.9\% \\
30 & 182.1 $\pm$ 135.5 & 69.2\% \\
29 & 185.3 $\pm$ 130.5 & 70.4\% \\
\bottomrule
\end{tabular}
\caption{Selected DACP steps for LunarLanderContinuous-v3}
\end{table}
\subsubsection{Observations}
The following observations can be derived from Figure~\ref{fig:performance}, which summarizes the reward--size trade-offs across all evaluated environments.
\begin{itemize}
    \item \textbf{General downward trend.} Across most environments, we observe a broadly monotonic decrease in reward as pruning progresses, reflecting the expected trade-off between interpretability and performance.
    \item \textbf{Reward drop-off point.} For all pruning algorithms and most environments, a distinct drop-off point emerges after which further pruning leads to a sharp decline in reward.
    \item \textbf{Occasional reward improvements.} In a few cases, such as \textit{Acrobot-v1} (see Figure~\ref{fig:performance}), pruning temporarily improves reward despite increasing interpretability, sometimes even above the original performance. This behavior could occur because of overfitted trees; If the overfitting branches are removed, the trees ability to generalize improves.
    \item \textbf{Small deviations between pruning algorithms.} While all pruning strategies follow a similar overall trend, minor deviations appear in the intermediate pruning stages. This leads to the conclusion that the complexity of a given environment influences prunability. 

\end{itemize}

In general, we observe that DACP performs superior over most environments leading to the conclusion, that for reinforcement learning tasks, structural pruning is not sufficient and backtracking algorithms that use mid-pruning reward based evaluation are beneficial.

\subsection{Limitations}
This work is limited by several factors. First, due to computational constrains, we decided to only distill policies to a max tree size of 1024. For decision trees that are grown to a significantly larger or smaller size, pruning results may differ.
Another limitation is the question wether or not the number of leaf nodes is a sufficient metric for interpretability; the actual contents might impact the interpretability significantly \citep{freitas2014}. For example, a larger tree could be perceived as more readable than a smaller one because of clearly understandable attributes \citep{freitas2014}.\\

\subsection{Future Work}
Future research could focus on validating interpretability through user studies to better understand which metrics truly correlate with human understanding, including those not examined in this work. In addition, the effects of other pruning strategies could be explored, for example by combining different strategies. Finally, pruned and visualized policies could be applied in critical environments where human verification and transparency are necessary. 

\section{Conclusion}
\label{sec:conclusion}
In this work, we introduce a modern framework for distilling and pruning reinforcement learning policies with the goal of improving interpretability for humans. We build upon the works of \cite{kohler2025interpretable} by using their distilling algorithm. We introduce a similar benchmark, adding the policy size in the form of leaf nodes as a metric, provide a way of visualizing policy trees to improve readability by humans, and explore the effects of pruning to gain interpretability.
Our pruning experiments show, that simplification can improve interpretability for reinforcement learning policies without immediate reward loss, though a trade-off emerges once complexity is reduced too far. In some cases, pruning even improved performance by reducing overfitting. Crucially, our method makes the simplification process itself transparent, which is an advantage over directly training smaller models.
This study nevertheless faces limitations. Interpretability was measured only through proxies, leaving open the question of how well these align with actual human cognitive accessibility. While our leaf-node-based proxy and visualizations try to capture interpretability, controlled user studies remain necessary to validate the true impact of pruning. Overall, pruning offers a promising step toward reinforcement learning policies that are both effective and understandable.

% achievements and limitations

% follow-up experiments

\section*{Conflict of Interest}
The authors have no conflicts of interest to declare that are relevant to the content of this article.

\bibliographystyle{abbrvnat}      % mathematics and physical sciences
\bibliography{references}   % name your BibTeX data base

\section{Appendix}
\label{appendix}

\begin{algorithm}[h]
\caption{DACP (high-level)}
\label{alg:heuristic_pruning}
\KwIn{Policy $\pi$, tolerance: $0 \leq \delta \leq 1$, stability $0 \leq \alpha \leq 1$}
\KwOut{Pruned policy $\pi'$}
$R_{Last} \gets R_{Base} \gets benchmark(\pi)$ \;
$reset \gets \text{False}$\;
$F \gets \emptyset$ \;

$k \gets \sqrt{|\pi|}$, rounded down to power of two\;

\While{true}{
    $\mathbf{C} \gets$ $counters(\pi)$\;
    $R_{min} \gets \min\big(R_{\text{base}} - |R_{\text{base}}|(1-\alpha),\; R_{\text{last}} - |R_{\text{base}}|\delta\big)$\;
    $N \gets k$ least visited nodes from $\mathbf{C} \setminus F$\;

    \If{$N = \emptyset$}{
        \If{$reset$}{break\;}
        $F \gets \emptyset$\;
        $reset \gets \text{True}$\;
        $k \gets \sqrt{|\pi|}$, rounded down to power of two\;
        continue\;
    }

    $(\pi', R', F_ {new}, ok) \gets$ \textbf{PruneBatch}($\pi, N, R_{Last}, R_{min}$)\;

    \If{$|\pi| = |\pi'|$}{
        $F \gets F \cup N$\;
        continue\;
    }
    \Else{
        $F \gets F \cup F_{new}$\;
        $\pi \gets \pi'$\;
        $R_{Last} \gets R'$\;
        $reset \gets \text{False}$\;
        $\pi \gets \text{collaps}(\pi')$\;

        \If{\textbf{not} $ok$}{
            $k \gets \sqrt{|\pi|- |F|}$, rounded down to power of two\;
        }
        
    }
}
\Return{$\pi$}\;
\end{algorithm}

\begin{algorithm}[h]
\caption{PruneBatch}
\label{alg:pruneBatch}
\KwIn{Policy $\pi$, candidate nodes $N$, last reward $R$, minimum reward $R_{min}$}
\KwOut{New policy $\pi'$, new reward $R'$, failed nodes $F$, success flag $ok$}

$\pi' \gets$ remove $N$ from $\pi$\;
$R'\gets benchmark(\pi')$\;

\If{$R' \geq R_{min}$}{ \Return{$\pi', R', \emptyset, True$}\;}
\If{$N = \emptyset$}{\Return{$\pi, R, \emptyset, False$}\;}

\If{$|N| = 1$}{\Return{$\pi, R, N, False$}\;}
$N_L, N_R \gets$ split($N$)\;
$(\pi_L, -, F_L, -) \gets$ \textbf{PruneBatch}($\pi, N_L, R, R_{min}$)\;
$(\pi_R, R_R, F_R, -) \gets$ \textbf{PruneBatch}($\pi_L, N_R, R, R_{min}$)\;

\Return{$\pi_R, R_R, F_L \cup F_R, False$}\;

\end{algorithm}

%\begin{acknowledgements}
%\end{acknowledgements}

%\bibliographystyle{spmpsci}      % mathematics and 

\end{document}